\documentclass{article}
\usepackage{booktabs}
\usepackage{float}
\usepackage{amsmath}
\restylefloat{table}
\usepackage{graphicx}
\usepackage{authblk}
\usepackage[margin=1in]{geometry}

\title{Quantaized Winograd/Toom-Cook Convolution for DNNs:\\ Beyond Canonical Polynomials Base}
\date{}	
\author{Barbara Barabasz}
\affil{School of Computer Science and Statistics, \\
		Trinity College Dublin,
		Ireland}
	
\begin{document}
	\maketitle	
\begin{abstract}
	The problem how to speed up the convolution computations in Deep Neural Networks is widely investigated in recent years. The Winograd convolution algorithm is a common used method that significantly reduces time consumption. However, it suffers from a problem with numerical accuracy particularly for lower precisions. In this paper we present the application of base change technique for quantized Winograd-aware training model. We show that we can train the $8$ bit quantized network to nearly the same accuracy (up to 0.5\% loss) for tested network (Resnet18) and dataset (CIFAR10) as for quantized direct convolution with few additional operations in pre/post transformations. Keeping Hadamard product on $9$ bits allow us to obtain the same accuracy as for direct convolution.   
\end{abstract}

\section{Motivation}
\paragraph{}
	The two-dimensional direct convolution algorithm has complexity of $O(k^2)$ multiplications to compute a single output point for kernel of the size $k\times k$. Using Winograd method we need $(o+k-1)^2$ multiplications to compute $o^2$ output points. That gives us linear complexity instead of square one. Also the bigger output $o$ we use the less multiplications we need to convolve all input. In practice, speedups up to $4\times$ can be achieved \cite{Maji19}.

The idea of Winograd algorithm is to transform inputs and weights into other domain, where convolution becames elementwise multiplication and then transform the results back. As we use the same kernel and inputs several times the cost of transformations amortizes over multiple uses. The main problem that transformations bring, is the loss of the accuracy of the final results. 

As the transformation matrices (Vandermonde matrices) have a bad properties in real field \cite{Pan16} the numerical error increase at least exponentially with the output size. It makes it necessary to break input into smaller tiles $(4\times 4, 6\times 6)$. Even then we could get a huge accuracy loss when considering lower precisions like $8$ bits.

\section{Literature}
\paragraph{}
There is not a lot of work done about Winograd algorithm in $8$ bits. Meng and Brothers in \cite{Meng19} proposed quantized Winograd algorithm with polynomials $x$, $x+1$, $x-1$, $x^2+1$ (the general version of the algorithm with superlinear polynomials is described in details in \cite{Barabasz19}. However, this method increases the number of general multiplications from $2.25$ to $3.06$ for single output point (two-dimensional convolution with kernel $3\times 3$ and output $4\times 4$). It also increase the number of dot product computations in pre/post transformation operations, where used matrices with one row/column more than in optimal Toom-Cook version. Their approach gives reasonable balance between number of real multiplications and the accuracy of the result. 
In this paper we propose the method that consists of optimal number of general multiplication operations, keep all stages of Winograd convolution in $8$ bits and give nearly the same image recognition accuracy (for tested network) as direct convolution at the expense of additional matrix multiplications in pre/post transformations. While keeping Hadamard product result on $9$ bits, we can train the network up to the same accuracy of image recognition as with $8$ bits direct convolution.

In \cite{Fernandez20}  Fernandez-Marques at al. proposed Winograd-aware training with symmetric $8$ bits quantization. The idea bases on including Winograd algorithm in convolution layers instead of direct one during training. Two methods are presented: first keep the transformation matrices ($G$, $B^T$ and $A^T$) fixed (\textit{static}) and including Winograd algorithm in convolution layers with transformation matrices as learnable parameters (\textit{flex}). They got a very good results for Winograd(Toom-Cook) algorithm with output $2 \times 2$, but they observe a loss in accuracy of image recognition for version with output $4 \times 4$ and $6 \times 6$. In all considered cases they got a huge improvement comparing to results got while train network with direct convolution layer and then replace it with Winograd/Toom-Cook algorithm. It suggests that the idea have a big potential for further investigation and improvement and we use it in this paper.

We perform the transformations in Winograd algorithm in Legendre polynomial base instead of canonical one. It requires additional operation in pre/post transformations to change the base but keeps the optimal number of general multiplications and improves the final accuracy.

\section{Winograd algorithm}
\begin{figure}[h]
	\centering
	\includegraphics[scale=0.7]{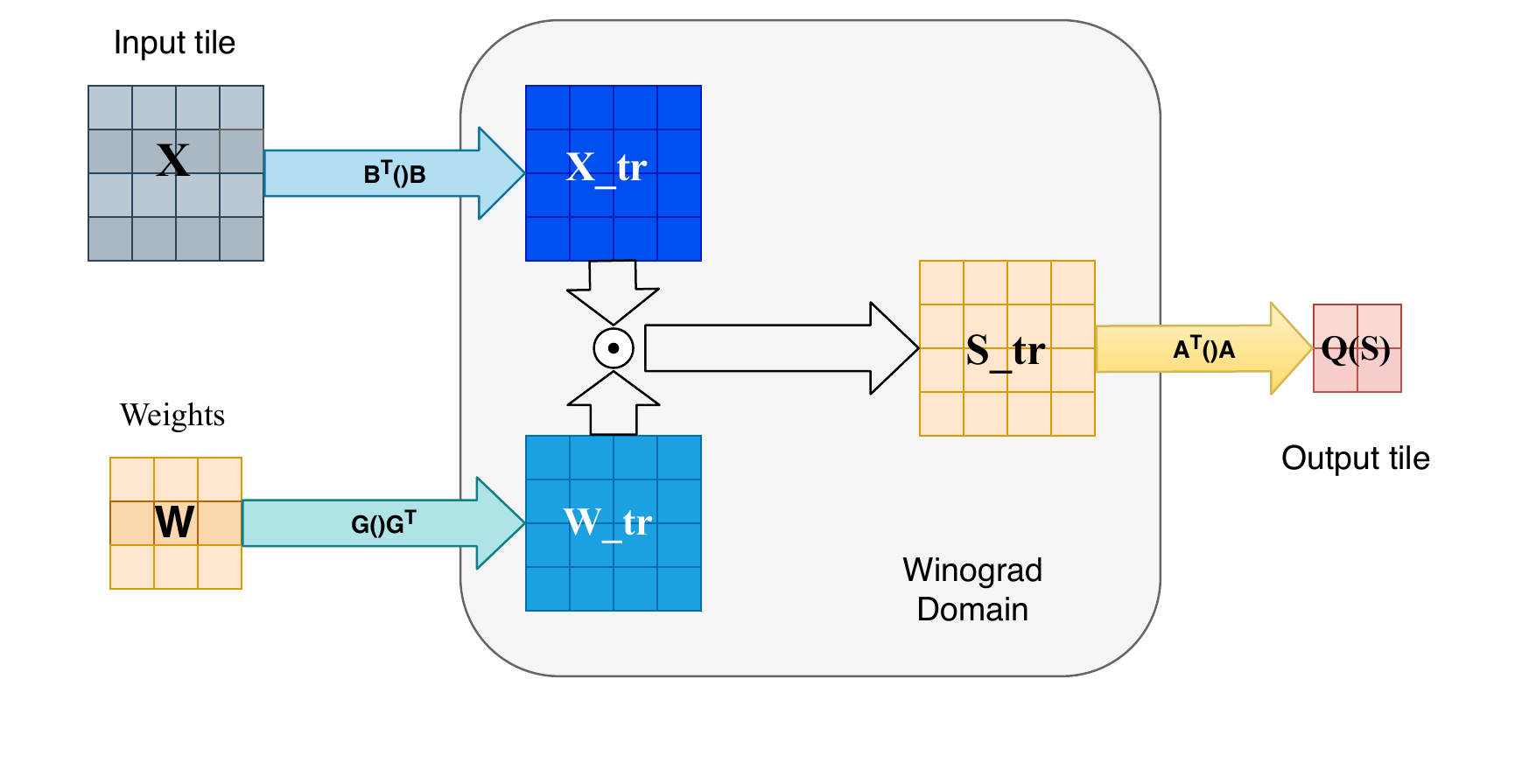}
	\caption{Winograd convolution algorithm}
	\label{fig:Win}
\end{figure}

\paragraph{}
The Winograd algorithm is based on Chinese Reminder Theorem (CRT) for polynomials and Matrix Exchange Theorem. The simplest version of this algorithm was presented by Toom \cite{Toom63}, Cook \cite{Cook66}. In this algorithm there are only linear polynomials used in CRT and the problem is equivalent to the interpolation problem. In $80's$ Schmuel Winograd has prooven the optimality of Toom-Cook algorithm according to the number of general multiplications (number of elementwise multiplications) and applied it to the signal processing \cite{Winograd80},\cite{Winograd80b}. He also proposed the more general version using polynomials of the degree higher than $1$ in CRT.  The goal of the Winograd algorithm is to transform inputs and weights into the 'Winograd domain' where convolution became an elementwise multiplication (Hadamard product) and then transform the result back. We need to construct three matrices: $G$, $B$ and $A$ used to perform transformation of weights, input and convolution result respectively (see figure \ref{fig:Win}).
More detailed descriptions of this construction could be find in \cite{Blahut10}, \cite{Tolimeri97}, \cite{Barabasz18}.

\section{Methodology} 
\label{sec:methodology}
\subsection{Transformation in Legendre polynomials base}
\paragraph{}
One of the reason that the Winograd convolution algorithm give less accurate results that direct convolution algorithm is the property of transformation matrices (Vandermonde matrices). In real field they are ill-posed (the condition number is big, so they are very close to singular matrices) and the error increase at least exponentially with their size \cite{Pan16}. The common method to decrease the condition number of this kind of matrices is to change the base. By default we use canonical polynomial base that means $1,x^2,x^3,x^4,\ldots$. The most popular bases that improve the numerical accuracy of such evaluation are Legendre, Czebyszew or Hermite polynomials. We implemented our idea using "normalised" Legendre polynomials that means the coefficient stands for the highest degree is equal to $1$. 

The disadventage of this method is bigger number of operations while performing transformations. For $2$ dimensional convolution, weights transformation includes two matrix multiplications:
\begin{equation}
\label{eq:weights_transf}
Weights\_transformed = G \thinspace W \thinspace G^T
\end{equation}
where $G$ is Vandermonde transformation matrix computed time ahead and $W$ stands for weights. 
If we assume that matrix $P$ stands for base change we need to perform $4$ matrix multiplications instead of $2$:
\begin{equation}
\label{eq:weights_transf_base}
Weights\_transformed = P^{-1} \thinspace G_P \thinspace W \thinspace G_P^T \thinspace P^{-T}
\end{equation}
where $G_P=P \thinspace G$ is new transformation matrix in new base $P^{-1}$ is the matrix that change the result back into the canonical base. They both are computed ahead of time. Matrix $W$ stands for weights.

While the Winograd convolution computations has following form:
\begin{equation}
\label{eq:conv}
A^T \thinspace \left( \left( G \thinspace W \thinspace G^T\right) \odot \left( B^T \thinspace X \thinspace B\right) \right) \thinspace A
\end{equation}
Our method could be described by following equation:
\begin{equation}
\label{eq:conv_base}
A_P^T \thinspace \left[P^{-T} \thinspace \left[ \left( P^{-1} \thinspace \left(G_P \thinspace W \thinspace G_P^T \right)\thinspace P^{-T}\right) \odot \left(B_P^T \thinspace \left(P^{-T} \thinspace X \thinspace P \right)\thinspace B_P\right) \right] \thinspace P^{-1}\right] \thinspace A_P
\end{equation}
where $X$ and $W$ stands for input tile and weights respectively. Matrices are computed in following way: $G_P = P \thinspace G$, $B_P = P \thinspace B$, $A_P = P \thinspace A$ ($G$,$B$ and $A$ are matrices constructed via Winograd(Toom-Cook) algorithm) and $P$ stands for base change matrix.

The larger number of operations, however, are in transformations, so amortizes over multiple uses of the same weights and input tiles. The number of general multiplications in Hadamard product (element-wise multiplications) is still the same as for Winograd(Toom-Cook) algorithm, that means it is optimal for given output/kernel sizes. It is also important to notice that matrix $P$ is sparse in particulary for smaller sizes. The matrices of the size $4\times 4$ and $6 \times 6$ include $6$ and $12$ non zero elements, respectively. 

In this paper we use the 'normalised' matrices in following forms:
$$
P^T = \begin{bmatrix} 1 & 0 & 0 & 0 & 0 & 0 \\
                    0 & 1 & 0 & 0 & 0 & 0 \\
                    -\frac{1}{3} & 0 & 1 & 0 & 0 & 0 \\
                    0 & -\frac{3}{5} & 0 & 1 & 0 & 0 \\
                    \frac{3}{35} & 0 & -\frac{6}{7} & 0 & 1 & 0 \\
                    0 & \frac{5}{21} & 0 & -\frac{10}{9} & 0 & 1 
 \end{bmatrix}
\qquad
P^{-T} = \begin{bmatrix} 1 & 0 & 0 & 0 & 0 & 0 \\
0 & 1 & 0 & 0 & 0 & 0 \\
\frac{1}{3} & 0 & 1 & 0 & 0 & 0 \\
0 & \frac{3}{5} & 0 & 1 & 0 & 0 \\
\frac{1}{5} & 0 & \frac{6}{7} & 0 & 1 & 0 \\
0 & \frac{3}{7} & 0 & \frac{10}{9} & 0 & 1 
\end{bmatrix}
$$

\subsection{Winograd-aware quantized training}
\begin{figure}[h]
	\centering
	\includegraphics[scale=0.7]{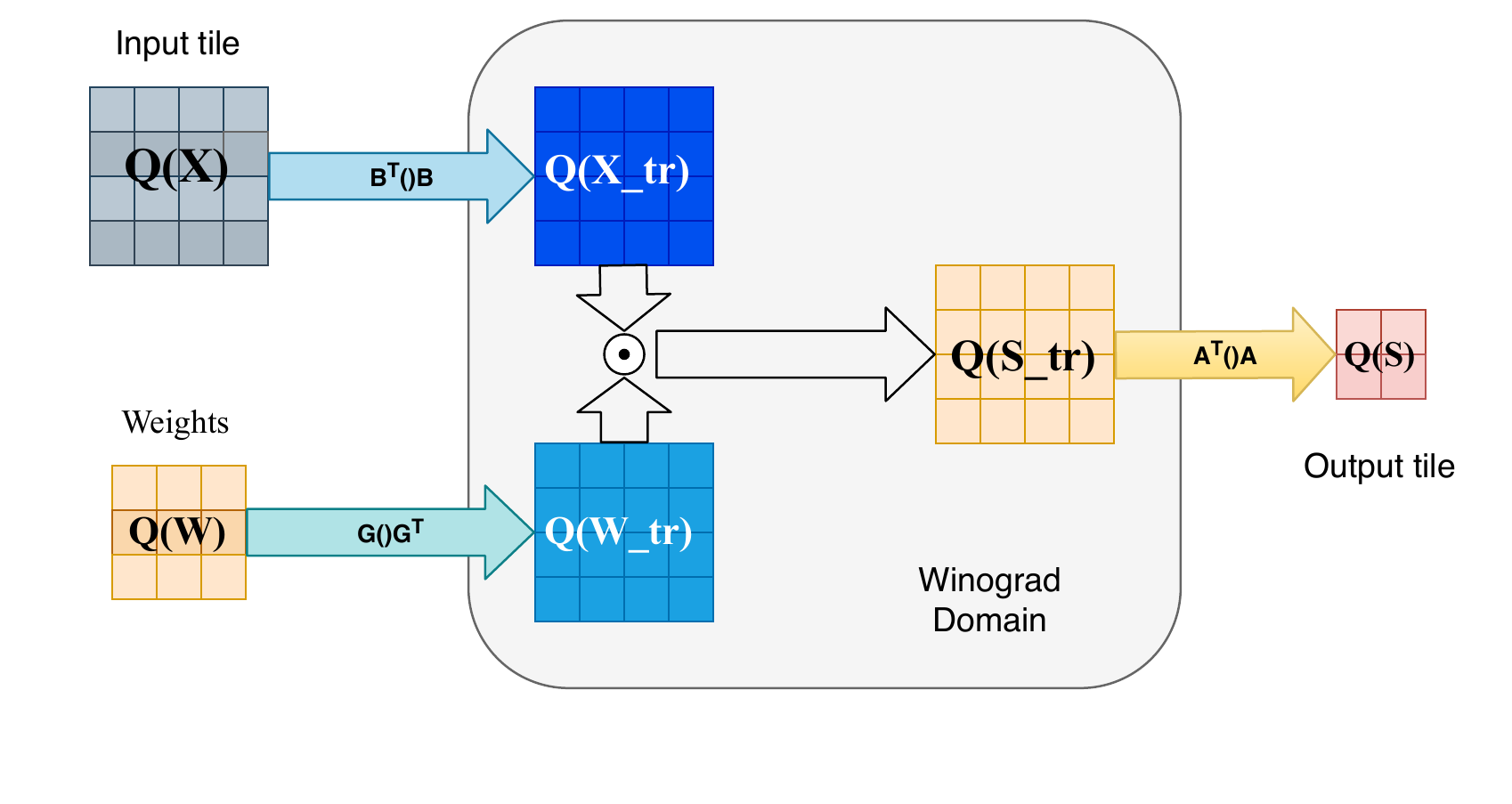}
	\caption{Quantized Winograd convolution algorithm}
	\label{fig:Win_Q}
\end{figure}

\paragraph{}
We merge the approach described in section \ref{sec:methodology} with Winograde-aware training method for $8$ bits quantization \cite{Fernandez20}. The authors tested two version \textit{static} - where transformation matrices $(G,B^T,A^T)$ are fixed and \textit{flex} - where transformation matrices are trainable parameters. In our test in \textit{flex} version we treat metrices $G_P, A_P, B_P$ as trainable parameters and leave $P$ and $P^{-1}$ fixed. Then we do not increase the number of trained parameters.  
The $8$ bits symmetrical quantization was implemented as casting the values before and after all transformations as presented on figure \ref{fig:Win_Q}

\section{Results} 
\paragraph{}
We run some preliminary experiments based on open source code \footnote{https://github.com/jafermarq/WinogradAwareNets} \cite{Fernandez20}. In this paper we present the results for Resnet18 network with channel multiplication factor equal to $0.25$ and $0.5$ with CIFAR10 database. We replace the existing Winograd layer with the new one that include the proposed transformations of input, weights and output. We consider Winograd(Toom-Cook) algorithm with kernel of the size $3\times 3$ and output $4 \times 4$. We denote as $L$ the version of Winograd convolution algorithm in Legendre polynomials base performed according to the formula \ref{eq:conv_base} with matrix $P$ as presented in section \ref{sec:methodology}.

	\begin{table}[h]
		\centering
    \caption{Resnet18 with number of channels coefficient equal to $0.5$, Winograd F4 in canonical and Legendre polynomials base, $8$ bits and $8$ bits with $9$ bits for Hadamard product.}
	\begin{tabular}{lccccc}
		\toprule 
		& direct & Static & Flex & L - static & L - flex \\ 
		\midrule

    8 bits & \textbf{92.3\%} & 77.2\% & 91.1\% & 85.0\% & \textbf{91.8\%} \\
    8b + 9b & - & 78.2\% & 91.5\% & 89.4\% & \textbf{92.3\%}\\
	\bottomrule 
	\end{tabular}
   \end{table}

\begin{table}[h]
	\centering 
	\caption{Resnet18 with number of channels coefficients equal to $0.25$ and $0.5$, $8$ bits quantization, Winograd F4 in canonical and Legendre polynomials base for transformations.}
	\begin{tabular}{lccccc} \toprule
	    mult & direct & Static & Flex & L - static & L - flex \\
	    \midrule 
	    $0.25$ & \textbf{90.2\%} & 74.0\% & 89.1\% & 81.9\% & 89.7\% \\
	    $0.5$ & \textbf{92.3\%} & 77.2\% &  91.1\% & 85.0\% & 91.8\% \\
	    \bottomrule
	\end{tabular}
\end{table}

We found that changing the base of transformation matrices has a significant impact on the accuracy of the quantized Winograd-aware training. 
The results for $8$ bits quantization \textit{static} version are still not satisfactory, but the improvement is over $7\%$ and show the potential of this method. For \textit{flex} version the error is reduced by more than half and give only $0.5\%$ worse accuracy comparing to the direct convolution, keeping the optimal number of general multiplications. 
In presented results for $8$ bits quantization with $9$ bits for Hadamard product we have the same accuracy as for direct convolution. It suggest that in future work we should focus on improvement of this stage of Winograd algorithm. Transformations accuracy on $8$ bits in Legendre polynomial base for output $4 \times 4$ do not require further improvement.

\section{Conclusion}
\paragraph{}
In this paper we present the idea for improving the Winograd-aware quantized networks. It required only a few additional operations in pre/post transformations and keep the optimal number of general multiplications. The first tests give the promising results. 

We show that changing the base of transformations into Legendre polynomials base for  $8$ bits quantized \textit{flex} Winograd-aware training we reduce the error by more than half comparing to the  quantized Winograd-aware training in canonical base. We obtain only $0.5\%$ worse accuracy of image recognition than for direct convolution (Resnet18, CIFAR10). While use $9$ bits for Hadamard product we have the same image recognition accuracy as for direct convolution. 

We show that in our tests the accuracy of $8$ bits quantization transformations are good enough and the reason of the acuracy loss lie in Hadamard product computations. By increasing the number of bits on this stage to $9$ we fully close the gap between Winograd/Toom-Cook algorithm with kernel $3\times 3$ and output $4\times 4$ and direct convolution in tested cases. For $8$ bits quantized Winograd-aware training with transformations in canonical base and Hadamard product on $9$ bits we still have $0.8\%$ loss in accuracy.

We plan to perform further tests for presented algorithm with other networks and datasets. We expect that networks with bigger number of channels or images size might required more bits for Hadamard product stage.

\section*{Acknowledgements}
This work was supported by Science Foundation Ireland grant 12/IA/1381. I would like to thank the grant holder Prof. David Gregg. I also extend my thanks to Javier Fernandez-Marques from Department of Computer Science, Oxford University for discussion.

\bibliographystyle{plain}
\bibliography{references_L}

\end{document}